\documentclass[10pt,leqno]{amsart}
\usepackage{graphicx} 
\usepackage{indentfirst,csquotes}
\usepackage{subcaption}
\usepackage{placeins}
\usepackage{amssymb,amsthm,amsmath}
\usepackage{xcolor,paralist,hyperref,fancyhdr,etoolbox}

\hypersetup{ colorlinks=true, linkcolor=black, filecolor=black, urlcolor=black }

\usepackage{lipsum}

\begin{document}
\title{Generating Synthetic Invoices via Layout-Preserving Content Replacement} 
\author[Initial Surname]{Bevin V, Ananthakrishnan P V, Ragesh KR, Sanjay M, Vineeth S, Bibin Wilson \\
\textbf{BEO AI}}
\thanks{The code for this project is available at: \url{https://github.com/BevinV/Synthetic_Invoice_Generation}}
\address{Address}
\email{bevivino789@gmail.com}
\maketitle

\begin{abstract}
The performance of machine learning models for automated invoice processing is critically dependent on large-scale, diverse datasets. However, the acquisition of such datasets is often constrained by privacy regulations and the high cost of manual annotation. To address this, we present a novel pipeline for generating high-fidelity, synthetic invoice documents and their corresponding structured data. Our method first utilizes Optical Character Recognition (OCR) to extract the text content and precise spatial layout from a source invoice. Select data fields are then replaced with contextually realistic, synthetic content generated by a large language model (LLM). Finally, we employ an inpainting technique to erase the original text from the image and render the new, synthetic text in its place, preserving the exact layout and font characteristics. This process yields a pair of outputs: a visually realistic new invoice image and a perfectly aligned structured data file (JSON) reflecting the synthetic content. Our approach provides a scalable and automated solution to amplify small, private datasets, enabling the creation of large, varied corpora for training more robust and accurate document intelligence models.
\end{abstract} 

\section{Introduction}

The digital transformation of financial workflows has made the automated processing of invoices a cornerstone of modern business operations. Machine learning (ML), particularly deep learning models, has shown significant promise in accurately extracting structured data from these documents, thereby reducing manual labor, minimizing errors, and accelerating payment cycles.

However, the performance of these sophisticated models is fundamentally bottlenecked by the availability of large-scale, diverse, and meticulously labeled training data. The acquisition of such datasets is fraught with challenges. Real-world invoices contain sensitive financial and personal information, making them subject to strict privacy regulations like GDPR. Furthermore, the manual process of collecting, scanning, and annotating thousands of invoices is prohibitively expensive and time-consuming.

To overcome this data scarcity bottleneck, we propose a novel pipeline for generating high-fidelity, synthetic invoice documents paired with their corresponding structured data. Our method leverages a hybrid approach, combining the strengths of Optical Character Recognition (OCR) for layout preservation, Large Language Models (LLMs) for contextual data generation, and computer vision techniques for seamless image composition. The system first deconstructs a source invoice into its textual and spatial components, then intelligently replaces key information with realistic, synthetic content, and finally re-renders a new, visually coherent invoice.

The key contributions of this work are: 
\begin{enumerate}
    \item A complete, end-to-end pipeline for generating paired image-JSON invoice data.
    \item A method for leveraging LLMs to produce contextually aware, anonymized data that maintains document plausibility.
    \item A robust text-rendering technique that preserves the original document's layout and font scale.
\end{enumerate}
This paper will first review related work in synthetic data generation, then detail our methodology, present the results of our generation process, and finally discuss the implications and future directions of this research.

\section{Related Work}

The creation of training data for document intelligence models has evolved significantly. We categorize prior work into three main areas: classical augmentation, generative modeling, and layout-aware language models.

\subsection{Classical and Content-Agnostic Augmentation}
The foundational approach to expanding datasets is data augmentation. For image-based tasks, this involves applying transformations that preserve class labels. A comprehensive survey by Shorten and Khoshgoftaar \cite{Shorten2019} details common techniques like rotation, scaling, noise injection, and elastic distortions. While effective for making models robust to visual noise and scanning artifacts, these methods are content-agnostic; they do not alter the semantic information within the document. More advanced techniques like "Cutout" \cite{DeVries2017}, which randomly erases rectangular regions of an image, force models to learn from less complete data but still do not modify the underlying text. These methods fail to enrich the variety of textual entities (e.g., company names, line items) that a document extraction model must learn.

\subsection{Generative Models for Document Synthesis}
To generate novel content, researchers have turned to generative models, most notably Generative Adversarial Networks (GANs). Early applications focused on generating realistic images, with seminal work by Goodfellow et al. \cite{Goodfellow2014} laying the groundwork. Subsequent work has attempted to adapt GANs for document synthesis. For example, LayoutGAN \cite{LayoutGAN2019} generates layouts by modeling the spatial relationships between document elements, but it does not generate the textual content itself. Conversely, models like MC-GAN \cite{MCGAN2017} can generate images conditioned on text, but often struggle with the legibility and precise placement required for complex documents. The primary drawback of most GAN-based approaches for document AI is the difficulty in producing coherent, readable text and ensuring that the visual output perfectly matches any corresponding structured data label.

Another line of work involves programmatic generation. The work by C. L. Tan et al.\cite{Tan2002} shows an early approach using templates and rendering engines to create synthetic documents for OCR training. While this guarantees perfect data labels, such template-based systems often produce datasets with limited layout diversity, which may not generalize well to the variance of real-world documents.

\subsection{Layout-Aware Language Models}
The most recent and successful approaches combine the power of Large Language Models (LLMs) with an understanding of document layout. The LayoutLM series from Microsoft represents a major breakthrough in this area. LayoutLM \cite{LayoutLM2020} was the first model to jointly pre-train text and layout information by incorporating 2D position embeddings. LayoutLMv2 \cite{LayoutLMv2_2021} improved upon this by adding visual feature embeddings directly from the image, creating a truly multimodal model. The culmination, LayoutLMv3 \cite{LayoutLMv3_2022}, simplified the architecture and used a unified text-image masking objective, setting a new state-of-the-art on many document AI benchmarks.

Following this trend, other models like "Donut" (Document Understanding Transformer) \cite{Donut2021} proposed an end-to-end approach that bypasses traditional OCR, using a Transformer encoder-decoder to directly generate a structured data output from a document image. While incredibly powerful for information extraction, these models are primarily discriminative; they are designed to *read* documents, not *generate* new ones.

The gap in the literature, therefore, lies in the generation of new, paired data. While generative models struggle with data-label consistency and layout-aware models are built for extraction, a practical pipeline for data *augmentation* through synthesis is still needed. Our work addresses this specific gap. By anchoring our generation process on the layout of a real document, we inherit its visual realism. By leveraging an LLM for its core competency—contextual text generation—and then re-rendering, we create a new, visually plausible document with a perfectly corresponding structured data label, a critical requirement for training the next generation of supervised document intelligence models.

\section{Methodology}

\begin{figure*}[htbp]
    \centering
    \includegraphics[width=0.5\textwidth]{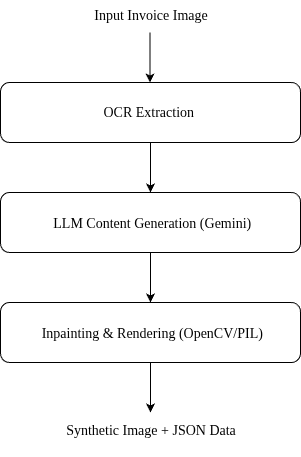}
    \caption{The end-to-end pipeline of our synthetic invoice generation system. The process begins with a source invoice, deconstructs it using OCR, generates new text with an LLM, and reconstructs a new, synthetic invoice image along with its corresponding structured JSON data.}
    \label{fig:pipeline}
\end{figure*}

To address the need for paired image and structured data, we developed a multi-stage pipeline that transforms a single seed invoice into a new, synthetic, and visually plausible counterpart. Our approach ensures that the layout and style of the original document are preserved while the textual content is replaced. The entire process, illustrated in Figure~\ref{fig:pipeline}, is composed of three core stages: OCR-based Content and Layout Extraction, LLM-Powered Content Generation, and finally, Image Inpainting and Text Rendering.

\subsection{OCR-based Content and Layout Extraction}
The foundational stage of our pipeline involves digitizing the source document's structure. For this task, we employ the \texttt{doctr} OCR engine, chosen for its strong performance in recognizing full lines of text within complex document layouts. The engine processes the input invoice image and outputs a list of all detected text fragments. For each fragment, we extract two critical pieces of information:
\begin{itemize}
    \item \textbf{Text Content:} The verbatim string of text recognized by the OCR engine.
    \item \textbf{Bounding Box:} A set of four (x, y) coordinates defining the precise rectangular area occupied by the text on the page.
\end{itemize}
This extraction process effectively transforms the visual document into a structured list of semantic elements and their corresponding spatial information, which serves as a scaffold for the subsequent generation stage. A unique ID is assigned to each text fragment for tracking throughout the pipeline.

\subsection{LLM-Powered Content Generation}
With the document's content and structure extracted, we leverage a Large Language Model (LLM) to generate new, contextually appropriate text. We utilize Google's Gemini 1.5 Flash model for this task due to its robust reasoning and response formatting capabilities.

A curated list of text fragments to be replaced (e.g., company name, addresses, line item descriptions, totals) is compiled. This list, containing the ID and original text of each fragment, is formatted into a prompt for the LLM. The prompt instructs the model to generate a realistic but fictional replacement for each item, maintaining the context (e.g., a date is replaced with a date, a currency amount with another currency amount). To ensure consistent and parsable output, the LLM is explicitly asked to return a single JSON object that maps each fragment's ID to its new, synthetic text string. This step effectively anonymizes the data while simultaneously augmenting it with novel content.

\subsection{Image Inpainting and Text Rendering}
The final stage reconstructs the invoice image with the new synthetic content. This is a two-step computer vision process:

\textbf{1. Text Removal via Inpainting:} First, the original text must be erased from the image. We generate a binary mask from the bounding boxes of all the text fields designated for replacement. This mask, which is white in the areas to be removed and black elsewhere, is fed into OpenCV's inpainting algorithm (\texttt{cv2.INPAINT\_TELEA}). This algorithm intelligently fills the masked regions based on the pixels from the surrounding area, effectively removing the text while preserving the background texture and lines of the invoice.

\textbf{2. Synthetic Text Rendering:} Next, the new text is rendered onto the inpainted image using the Python Imaging Library (PIL). For each text fragment, we retrieve its original bounding box and its new synthetic text from the LLM. A critical challenge here is ensuring the new text fits naturally into the space of the old text. To solve this, we implement a dynamic font sizing algorithm. The font size is initially set based on the height of the bounding box and is then iteratively reduced until the rendered width of the new text is smaller than the width of the box. The text is then rendered, centered vertically and horizontally within the original bounding box, to maintain the document's visual alignment and professional appearance.

The output of this final stage is the synthetic invoice image, which is then saved alongside the generated JSON data, resulting in a perfectly paired training example.

\section{Experiments and Results}

To validate the efficacy of our pipeline, we conducted a series of generation experiments aimed at evaluating the quality, realism, and diversity of the synthetic invoices produced. The primary goal is to demonstrate that our method can generate visually plausible documents with perfectly aligned, structured data.

\subsection{Implementation Details}
Our pipeline was implemented in Python. We utilized the \texttt{doctr} library for OCR, which is built on PyTorch. For Large Language Model interactions, we used the Google Generative AI SDK to interface with the Gemini 1.5 Flash model. The image manipulation and rendering stages were handled by OpenCV and Pillow (PIL). All experiments were run on a standard workstation, demonstrating the accessibility of our approach. The font used for rendering new text was the open-source Arial font to ensure common character coverage.

\subsection{Qualitative Results}
\begin{figure*}[ht]
    \centering
    \begin{subfigure}[b]{0.50\textwidth}
        \includegraphics[width=\textwidth]{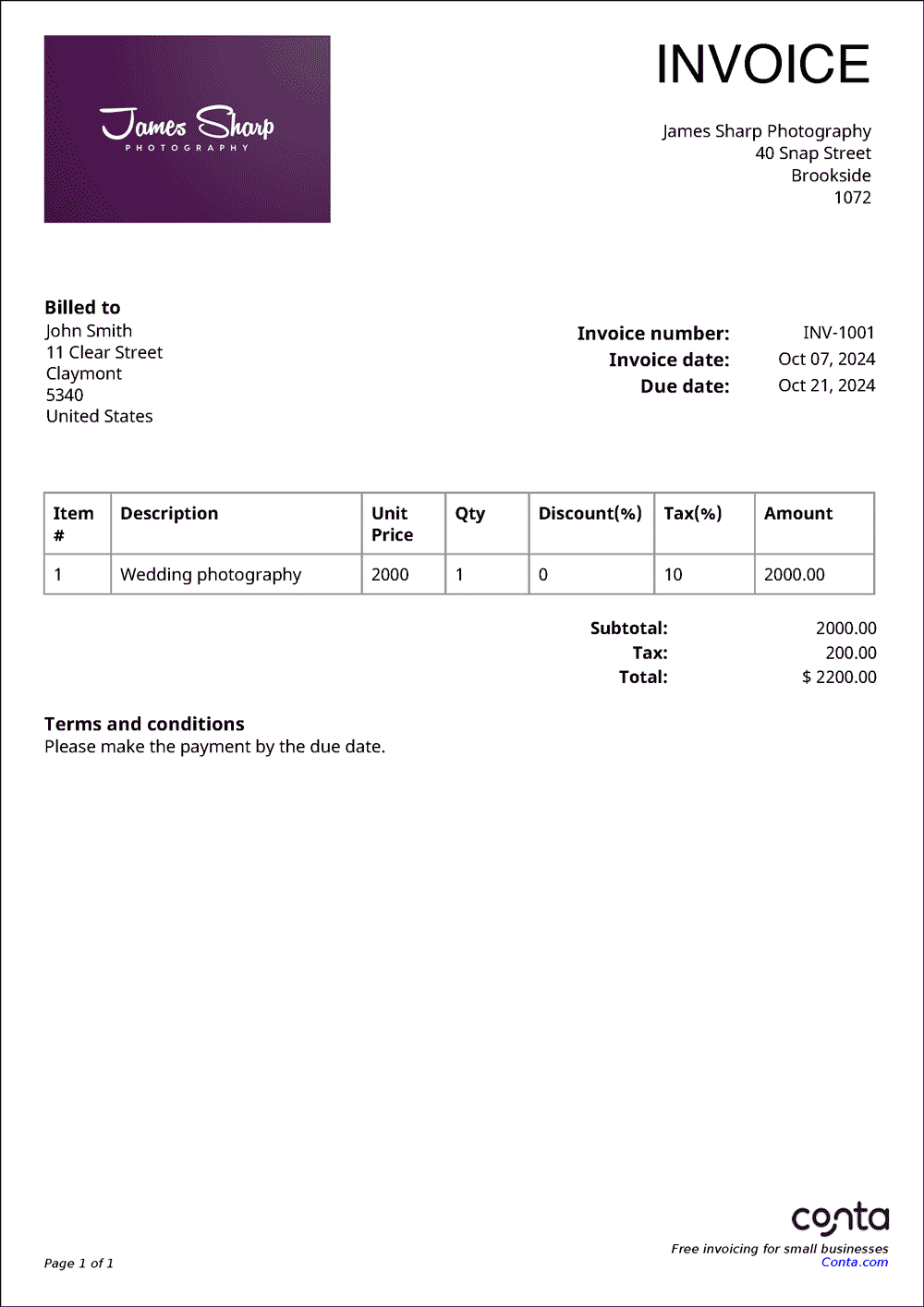}
        \caption{Original Source Invoice}
        \label{fig:comparison_a}
    \end{subfigure}
    \hfill 
    \begin{subfigure}[b]{0.48\textwidth}
        \includegraphics[width=\textwidth]{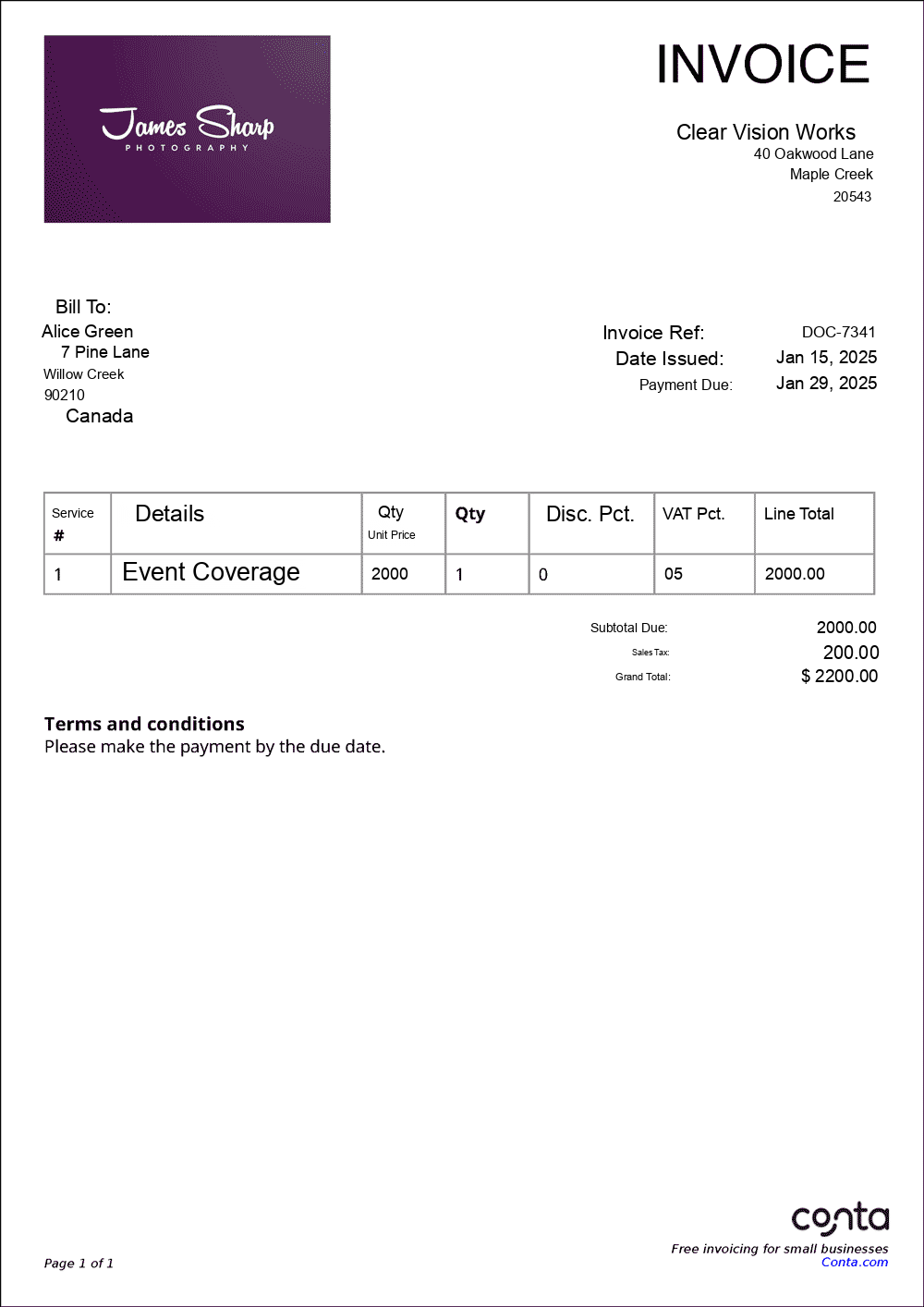}
        \caption{Synthetic Invoice Generated by our Pipeline}
        \label{fig:comparison_b}
    \end{subfigure}
    \caption{A detailed comparison between a source invoice and a synthetically generated output. The full-page width layout allows for inspection of fine details, showing that the layout is perfectly preserved while the textual content is replaced.}
    \label{fig:comparison}
\end{figure*}

The quality of synthetic data is often best assessed qualitatively. We present a detailed comparison between a source invoice and a synthetically generated output in Figure~\ref{fig:comparison}. This figure occupies the full page width to ensure legibility of the document's details.

As shown, the synthetic invoice (Figure~\ref{fig:comparison_b}) maintains the exact layout, background, and visual structure of the original (Figure~\ref{fig:comparison_a}). The original text has been seamlessly replaced with new, contextually plausible information generated by the LLM. Fields such as the recipient's name, address, invoice number, and line item descriptions are all novel, yet the document remains entirely coherent as an invoice. The high fidelity of the text rendering and inpainting is evident upon close inspection.

\subsection{Structured Data Correspondence}
A critical output of our pipeline is the structured JSON data that corresponds to the synthetic image. For every generated image, a JSON file is produced that mirrors the structure of the original invoice's data but is populated with the new synthetic values. For instance, if the original invoice's JSON was \texttt{\{"recipient\_name": "Original Name"...\}}, the new JSON will be \texttt{\{"recipient\_name": "Fictional Name"...\}}, where "Fictional Name" is the exact text rendered onto the new image. This guarantees a perfect one-to-one mapping between the visual document and its ground truth label, which is essential for training supervised machine learning models for document information extraction. This eliminates the possibility of data-label mismatch, a common source of error in manually annotated datasets.

\begin{figure*}[ht]
    \centering
    \begin{subfigure}[b]{0.48\textwidth}
        \includegraphics[width=\textwidth]{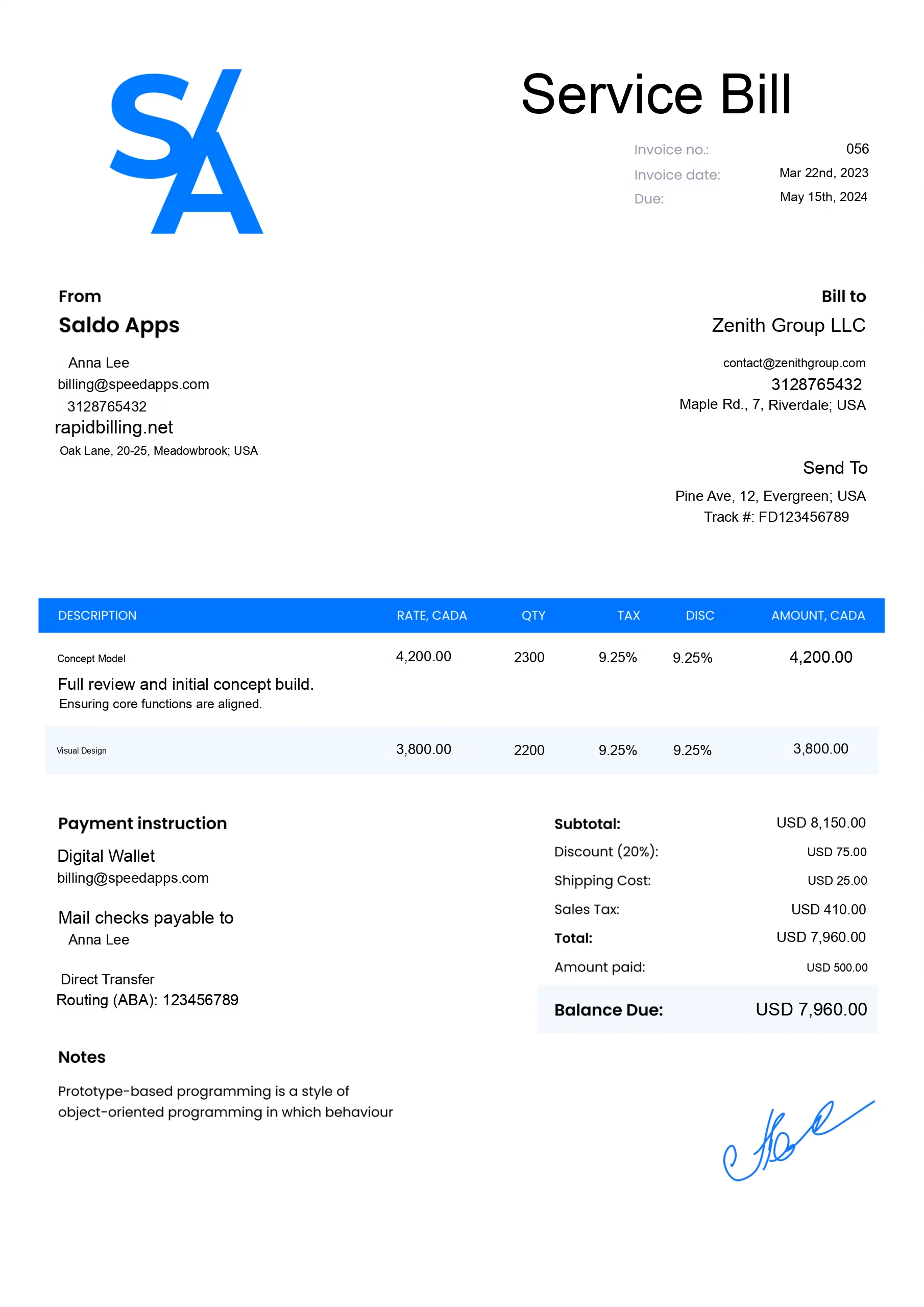}
        \caption{Synthetic Variant 1}
        \label{fig:variant_1}
    \end{subfigure}
    \hfill
    \begin{subfigure}[b]{0.48\textwidth}
        \includegraphics[width=\textwidth]{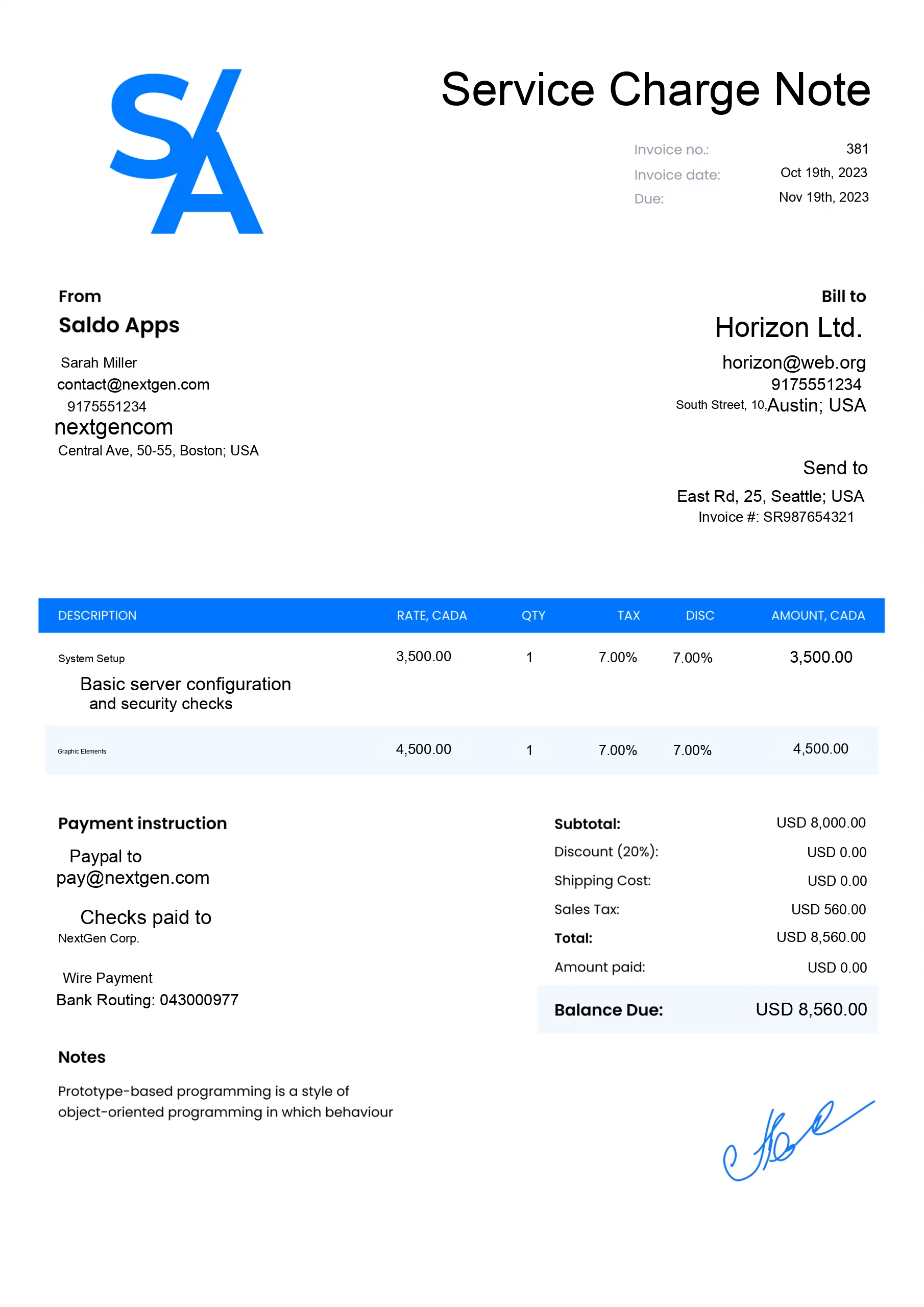}
        \caption{Synthetic Variant 2}
        \label{fig:variant_2}
    \end{subfigure}

    \vspace{0.5cm} 

    \begin{subfigure}[b]{0.48\textwidth}
        \includegraphics[width=\textwidth]{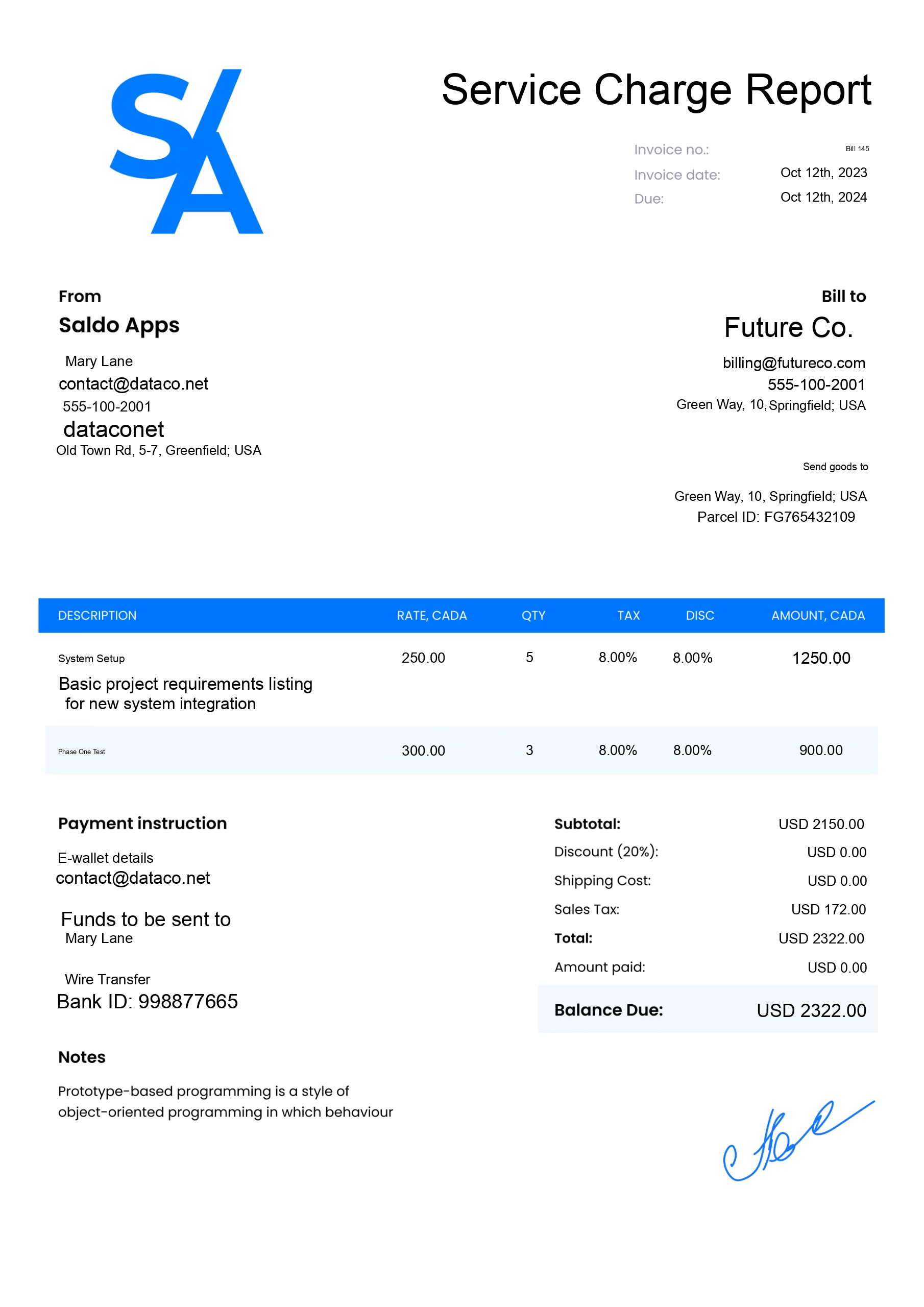}
        \caption{Synthetic Variant 3}
        \label{fig:variant_3}
    \end{subfigure}
    \hfill
    \begin{subfigure}[b]{0.48\textwidth}
        \includegraphics[width=\textwidth]{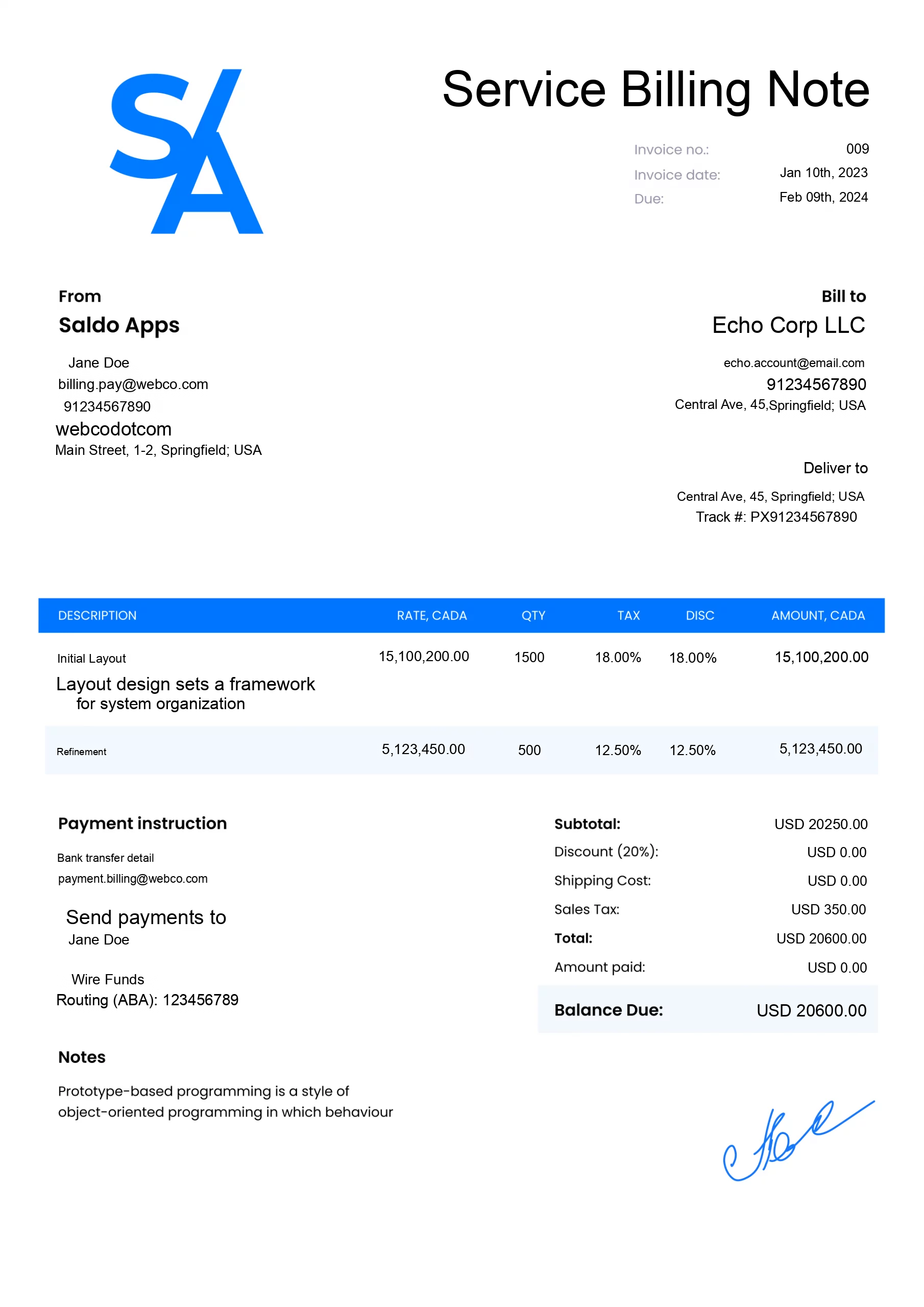}
        \caption{Synthetic Variant 4}
        \label{fig:variant_4}
    \end{subfigure}

    \caption{A showcase of generation diversity. All four synthetic invoices were generated from the same single source document. Note the variation in company names, addresses, dates, and line items, demonstrating the pipeline's ability to create a rich and varied dataset from a limited seed.}
    \label{fig:diversity}
\end{figure*}

To showcase the pipeline's ability to generate varied data from a single source, Figure~\ref{fig:diversity} presents multiple distinct outputs generated from the same seed invoice. Each example features unique company names, dates, financial figures, and other details. This demonstrates that our method can be used to amplify a small dataset by producing numerous unique variations, which is crucial for preventing a machine learning model from simply memorizing a few examples.

\newpage
\FloatBarrier

\section{Discussion and Future Work}

Our experiments demonstrate that the proposed pipeline is effective at generating realistic and diverse synthetic invoices. The key strength of our method lies in its hybrid nature: by anchoring the generation process on the layout of a real document, we preserve visual fidelity, while the use of a Large Language Model provides semantic plausibility and novelty. This approach directly addresses a critical need for high-quality, paired image-and-data examples for training document AI models.

\subsection{Limitations}
Despite its successes, our method has some limitations. The quality of the final output is heavily dependent on the quality of the initial OCR extraction. Errors in the OCR stage, such as missed text or incorrectly merged lines, can propagate through the pipeline. Secondly, while our dynamic font sizing works well, it does not attempt to precisely match the original font family, which could be a factor in some applications. Finally, the system does not currently generate non-textual elements like logos or stamps, which are common components of real-world invoices.

\subsection{Future Work}
There are several exciting avenues for future research. One direction is to improve the OCR-dependency by incorporating a feedback loop where the LLM could potentially correct OCR errors. Another promising area is the generation of non-textual elements by training a separate generative model (like a GAN or diffusion model) to create realistic logos that could be composited onto the synthetic invoices. Finally, we plan to conduct a quantitative study by training an extraction model on a dataset augmented by our method to formally measure the downstream performance improvement.

\section{Conclusion}

In this paper, we introduced SynthID, a novel and practical pipeline for generating synthetic invoices. By combining OCR, generative language models, and image inpainting, our method produces new, high-fidelity invoice images along with their perfectly corresponding structured data. We have shown that this approach can create a diverse set of training examples from a single seed document, providing a scalable solution to the data scarcity problem in document intelligence. This work enables researchers and developers to build more robust and accurate invoice processing models, even when access to large, real-world datasets is limited by privacy or cost.

\end{document}